\pdfoutput=1
\documentclass{article}
\usepackage{ijcai22}

\usepackage{times}

\usepackage{soul}
\usepackage{url}
\usepackage[hidelinks]{hyperref}
\usepackage[utf8]{inputenc}
\usepackage[small]{caption}
\usepackage{graphicx}
\usepackage{amsmath}
\usepackage{booktabs}
\usepackage{xparse}
\usepackage{algorithm}
\usepackage{algorithmicx}
\usepackage{algpseudocode}

\usepackage{amsmath}
\usepackage{booktabs,subcaption,amsfonts,dcolumn}
\usepackage{svg}
\usepackage{multirow}
\usepackage{bibentry}
\usepackage[american]{babel}
\usepackage{bm}

\makeatletter
\NewDocumentCommand{\LeftComment}{s m}{%
  \Statex \IfBooleanF{#1}{\hspace{85pt}}\(\triangleright\) #2}
\makeatother
\makeatletter
\NewDocumentCommand{\OneLineComment}{s m}{%
   \Statex  \IfBooleanF{#1}{\hspace{12pt}}\(\triangleright\) #2}
\makeatother
\makeatletter
\NewDocumentCommand{\LineComment}{s m}{%
  \Statex \IfBooleanF{#1}{\hspace{85pt}} #2}
\makeatother

\title{Neural Network Pruning by Cooperative Coevolution}

\author{
Haopu Shang$^1$\and
Jia-Liang Wu$^1$\and
Wenjing Hong$^2$\And
Chao Qian$^{1}$\thanks{This work was supported by the NSFC (62022039, 62106098), the Jiangsu NSF (BK20201247), and the CAAI-Huawei MindSpore Open Fund. Chao Qian is the corresponding author. The version without appendix is published on IJCAI-ECAI-22.} \\ 
\affiliations
$^1$State Key Laboratory for Novel Software Technology, Nanjing University, Nanjing 210023, China\\
$^2$Department of Computer Science and Engineering \\
Southern University of Science and Technology, Shenzhen 518055, China
\emails
\{shanghp, wujl, qianc\}@lamda.nju.edu.cn, hongwj@sustech.edu.cn
}

\begin{document}

\maketitle

\begin{abstract}
Neural network pruning is a popular model compression method which can significantly reduce the computing cost with negligible loss of accuracy. Recently, filters are often pruned directly by designing proper criteria or using auxiliary modules to measure their importance, which, however, requires expertise and trial-and-error. Due to the advantage of automation, pruning by evolutionary algorithms (EAs) has attracted much attention, but the performance is limited for deep neural networks as the search space can be quite large. In this paper, we propose a new filter pruning algorithm CCEP by cooperative coevolution, which prunes the filters in each layer by EAs separately. That is, CCEP reduces the pruning space by a divide-and-conquer strategy. The experiments show that CCEP can achieve a competitive performance with the state-of-the-art pruning methods, e.g., prune ResNet56 for $63.42\%$ FLOPs on CIFAR10 with $-0.24\%$ accuracy drop, and ResNet50 for $44.56\%$ FLOPs on ImageNet with $0.07\%$ accuracy drop.
\end{abstract}

\section{Introduction}

Convolution neural networks (CNNs) have achieved great success in computer vision tasks, such as image classification~\cite{DBLP:conf/cvpr/HeZRS16} and object recognition~\cite{DBLP:journals/ijcv/LiuOWFCLP20}. However, modern CNNs are typically too computationally intensive to be deployed on resource-constrained applications.
To address this issue, a variety of methods have been proposed to reduce their computational costs~\cite{DBLP:journals/corr/HintonVD15,DBLP:journals/corr/HanMD15}. Among them, neural network pruning has achieved impressive results in various applications by removing redundant parameters while keeping a high accuracy, as modern CNNs are usually overparameterized.

Despite its broad application prospects, the large number of parameters of CNNs poses a great challenge to neural network pruning. Performing the pruning process on the filter level can alleviate this issue to some extent. However, most previous filter pruning methods rely heavily on some criteria or auxiliary modules, which are used to measure the importance of each filter, but are usually derived from the experience and knowledge of experts. Thus, an automatic framework is essential to facilitate the application of filter pruning on various architectures and data sets.

Filter pruning can be generally formulated as a complex optimization problem with the aim of searching for a good subset of the original filters to be retained such that the resultant pruned network has a low computational cost while keeping a high accuracy. Evolutionary algorithms (EAs)~\cite{back1996evolutionary} are a kind of general-purpose heuristic randomized optimization algorithms, inspired by natural evolution. They have been naturally applied to solve the optimization problem of pruning, making the process automatic~\cite{DBLP:journals/ijis/Yao93}. However, unlike the multilayer perceptrons of earlier years, the large number of filters of modern CNNs implies a large search space, making EAs difficult to obtain a satisfactory solution within a limited computational overhead. As a consequence, existing EA-based pruning methods mainly take compromises when applied to modern CNNs. One way is to apply EAs to optimize the hyper-parameters (e.g., the pruning ratio of each layer~\cite{DBLP:conf/ijcai/LiQJLT18}) of existing pruning methods with expert knowledge. Others often use indirect encoding methods to compress the search space partly~\cite{2021Evolutionary,DBLP:journals/isci/FernandesY21}, which may, however, degrade the performance due to the loss of information.

To solve the essential large-scale neural network pruning problem, this paper proposes a novel Cooperative CoEvolution algorithm for Pruning (CCEP). Cooperative coevolution uses the idea of divide-and-conquer, and has been shown to be effective for solving large-scale optimization problems~\cite{DBLP:journals/tec/HongTZIY19}. Although cooperative coevolutionary EAs have been proposed before~\cite{DBLP:journals/tec/MaLZTLXZ19}, none of them has been designed with the explicit purpose to solve neural network pruning problems. The main idea of CCEP is to use the strength of cooperative coevolution to overcome the large-scale issue in neural network pruning, and meanwhile utilize the property of neural networks for the non-trivial separation issue of cooperative coevolution. Specifically, considering the characteristics of forward propagation that features are processed layer by layer, CCEP divides the original search space by layer because the impact of removing a filter is mainly related to whether the other filters in the same layer can retain the representation ability.

Experiments using ResNet and VGG on CIFAR-10 and ImageNet show that the proposed CCEP can achieve a competitive performance with a number of state-of-the-art pruning methods which require experts to design good criteria or auxiliary modules. Furthermore, compared with previous automatic EA-based pruning methods, CCEP can achieve both larger pruning ratio and smaller accuracy drop. Note that CCEP can be flexibly applied to different network architectures and data sets, since no complex rules or modules are employed and the optimization process is highly decoupled with the architecture. Due to the advantage of cooperative coevolution, CCEP can also be easily parallelized, which is friendly to the application in practice.

\section{Related Work}

\subsection{Neural Network Pruning}

Neural network pruning on CNNs has been widely studied in the past decade. The methods can be generally divided into non-structured and structured pruning, where the former is directly conducted on the weights and the latter is usually on the basic structured cells such as convolution filters. Nowadays, filter pruning has prevailed since it is friendly to most existing hardware and software frameworks.

Based on how to indicate the filters to be pruned, the mainstream filter pruning methods can be further categorized into criteria-based and learning-based methods. Criteria-based methods design some criteria artificially to identify unimportant filters and prune them. For example, L1~\cite{DBLP:conf/iclr/0022KDSG17} prunes the filters with smaller $l_1$-norm weights. FPGM~\cite{DBLP:conf/cvpr/HeLWHY19} calculates the geometric median of the filters in each layer and prunes the neighbouring filters. HRank~\cite{lin2020hrank} locates the low-rank features and prunes the corresponding filters. ThiNet~\cite{DBLP:journals/pami/LuoZZXWL19} prunes the filters by minimizing the reconstruction error. However, it is not easy to decide a proper criterion as well as the pruning ratio of each layer, which usually require rich experience and plenty of trials of experts. Learning-based methods use auxiliary modules to model the importance of filters in each layer and prune the neural network accordingly. For example, AutoPruner~\cite{DBLP:journals/pr/LuoW20} uses learnable indicators on filters as the reference to prune. BCNet~\cite{su2021bcnet} introduces a bilaterally coupled network to efficiently evaluate the performance of a pruned network, and performs the pruning based on this surrogate model. Such methods can liberate the trials on designing proper criteria, but usually lead to a complex pruning framework and deteriorate the extendibility.

\subsection{Pruning by Evolution}


Since the 1990s, EAs have been applied to prune artificial neural networks automatically~\cite{DBLP:journals/ijis/Yao93}, with the aim of removing unimportant neural cells and connections while keeping a high accuracy. As the scales of neural networks were small, EAs could be easily applied to achieve a good performance. However, in recent years, the architectures of neural networks have become more and more complex and large, hindering the direct application of EAs, since the large search space makes EAs hard to produce a good pruned network in acceptable running time.

The current EA-based pruning methods often search in a compressed filter space, e.g., DeepPruningES~\cite{DBLP:journals/isci/FernandesY21} shares an encoding in different layers, and ESNB~\cite{2021Evolutionary} uses block-level search. Instead of pruning neural networks directly, EAs are also applied to tune the hyper-parameters of existing pruning methods, e.g., OLMP~\cite{DBLP:conf/ijcai/LiQJLT18} searches a proper pruning threshold of each layer for layer-wise magnitude-based pruning methods. However, all these compromised strategies may lose useful information, leading to a less-satisfactory performance.

\subsection{Cooperative Coevolution}

Cooperative coevolution~\cite{DBLP:journals/ec/PotterJ00} is a famous evolutionary framework that has been shown suitable for large-scale optimization. It uses a divide-and-conquer approach to decompose a large-scale problem into several small-scale subcomponents and evolve these subcomponents cooperatively. A number of approaches have been proposed to incorporate the cooperative coevolution framework to improve the performance of EAs~\cite{DBLP:journals/tec/MaLZTLXZ19}. However, none of them is designed with the explicit goal of being able to solve neural network pruning problems. Furthermore, due to different properties of problems, it will often lead to a bad performance if applying a decomposition strategy designed for one problem directly to another. In this work, we adapt cooperative coevolution to neural network pruning for the first time, and employ a decomposition strategy based on the characteristics of forward propagation of CNNs.

\section{The CCEP Algorithm}

Let $\Phi$ denote a well-trained CNN with $n$ convolution layers $\{\mathcal{L}_1, \mathcal{L}_2,\ldots, \mathcal{L}_n\}$, where $\mathcal{L}_i$ indicates the $i$th layer and contains $l_i$ filters. The aim of neural network pruning is to determine a subset of filters in $\Phi$ for removal so that the resulting pruned network has maximum accuracy and minimum computational cost. Let $\bm{\sigma} = \{\sigma_{ij} \mid \sigma_{ij} \in \{0,1\}, i \in \{1,2,..., n\}, j \in \{1,2,...,l_i\}\}$, where $\sigma_{ij}$ denotes whether the $j$th filter (denoted as $\mathcal{L}_{ij}$) in the $i$th layer of the network $\Phi$ is retained. That is, if $\sigma_{ij}=1$, the filter $\mathcal{L}_{ij}$ is retained, otherwise it is removed. Thus, a vector $\bm{\sigma}$ naturally represents a pruned network $\Phi_{\bm{\sigma}} = \bigcup_{i=1}^{n} \bigcup_{j=1}^{l_i} \sigma_{ij}\mathcal{L}_{ij}$. When $\sigma_{ij}$ equals to 1 for any $i$ and $j$, $\Phi_{\bm{\sigma}}$ is actually the original network $\Phi$.

Under the solution representation based on vector $\bm{\sigma}$, neural network pruning can be formulated as an optimization problem that maximizes the accuracy and minimizes the computational cost of a pruned network $\Phi_{\bm{\sigma}}$ simultaneously. Using the number of FLoat point OPerations (FLOPs), which is a common metric, to measure the computational cost, the pruning problem can be formally described as
\begin{align}
    \mathop{\arg\max}\limits_{\bm{\sigma}\in \{0,1\}^{\sum^n_{i=1} l_i}} \mbox{    } & \big(\mbox{Accuracy}(\Phi_{\bm{\sigma}}),  -\mbox{FLOPs} (\Phi_{\bm{\sigma}})\big).
\label{problem}
\end{align}
This is essentially a large-scale optimization problem, as a deep CNN usually contains many filters, i.e., $\sum^n_{i=1} l_i$ is large. 

To solve the problem, a novel algorithm based on cooperative coevolution named CCEP, is proposed. The main idea is to make use of the divide-and-conquer technique applied in the cooperative coevolution framework, but transferring it to the context of neural network pruning. Despite cooperative coevolution has achieved impressive success in large-scale optimization, it has been reported that it would have a poor performance in non-separable problems if a proper decomposition strategy is not given~\cite{DBLP:journals/tec/MaLZTLXZ19}. As the filters in CNNs are usually highly related, the problem of neural network pruning is inherently non-separable. To solve this issue, a decomposition strategy based on the specific characteristics of typical CNNs is used in CCEP. Specifically, considering that the features are commonly processed layer by layer during the forward propagation of a CNN, the impact of removing a filter mainly depends on whether the other filters in the same layer can retain the representation ability. Thus, a natural strategy is to decompose the search space by layer. That is, the vector $\bm{\sigma}$ of decision variables is split into $n$ groups, where $\forall i \in \{1,2,..., n\}$, the $i$th group corresponds to the $i$th layer and contains $l_i$ variables $\{\sigma_{i1},\ldots, \sigma_{il_i}\}$. After that, CCEP coevolves $n$ subpopulations representing these groups, each of which is optimized with a separate EA. It is worthy noting that CCEP is highly decoupled with the architecture of the original network, and thus can be applied to different networks flexibly and extended easily. Details of the framework of CCEP and the EA in each group are described below.

\subsection{The Framework of CCEP}

\begin{figure}[tp]   \flushleft  \includegraphics[width=8.5cm]{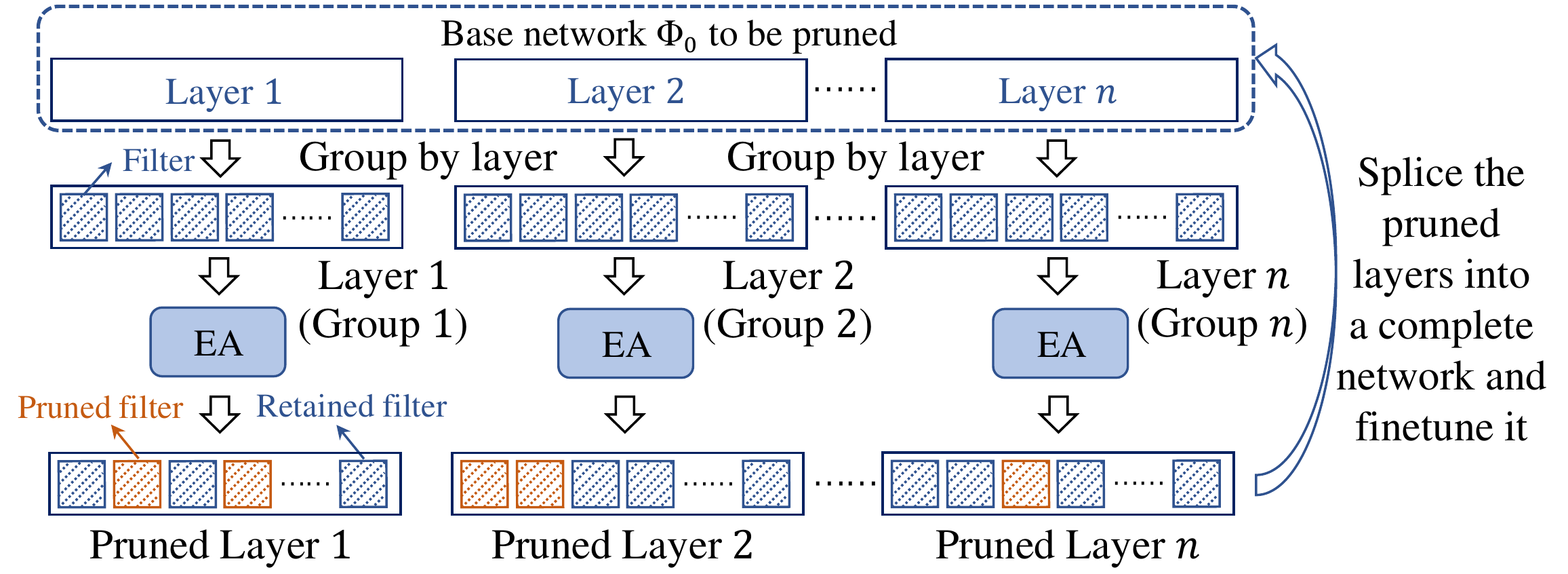} \vspace{-1.6em}
\caption{Illustration of the cooperative coevolution process of CCEP. A network with $n$ layers is iteratively evolved by being divided into $n$ groups, each of which represents a single layer, applying an EA in each group in parallel to obtain specific pruned layers, generating a complete pruned network by splicing the $n$ pruned layers, and finetuning the generated network.} \label{FIG:CCEP} \vspace{-1.3em} \end{figure}

The framework of CCEP is shown in Algorithm~\ref{alg:CCEP}. It employs an iterative prune-and-finetune strategy and finally outcomes a set of pruned networks with different pruning ratios for user selection. During each iteration, the algorithm works as follows. At the beginning, it splits the candidate filters to be pruned into $n$ groups by layer in line~4 of Algorithm~\ref{alg:CCEP}. Once the groups are created, the algorithm optimizes these groups in parallel in line~5. A separate EA is employed for each group, and it will continue until the stop condition specified in the EA is reached and outcome an individual representing the specific pruned layer. After that, a collaboration among these groups takes place in line~6, i.e., a complete pruned network will be created by splicing the $n$ resultant pruned layers in their intrinsic order. At last, the algorithm will perform a finetune process to recover the accuracy of the obtained complete pruned network in line~7. An illustration of one complete coevolution process is shown in Figure~\ref{FIG:CCEP}. The finetuned pruned network $\Phi'$ will be preserved into the archive $A$ in line~8, and used as the base network $\Phi_0$ in line~9 for the next iteration. After running $T$ iterations, the pruned networks in the archive $A$ are output in line~12.

\begin{algorithm}[tb]
\caption{CCEP framework}
\label{alg:CCEP}
\textbf{Input}: A well-trained CNN $\Phi$, maximum iteration $T$\\
\textbf{Output}: A set of pruned networks with different sizes\\
\textbf{Process}:
\begin{algorithmic}[1] 
\State Set the base network $\Phi_0 = \Phi$; 
\State Let $A = \emptyset$, and $i = 0$;
\While{$i$ \textless \  $T$}
\State Group the network $\Phi_0$ by layer;
\State Apply an EA in each group in parallel to obtain spe-
\Statex \quad \ \ cific pruned layers, as shown in Algorithm~\ref{FIG:EA};
\State Generate a complete pruned network by splicing all 
\Statex \quad \ \ pruned layers in their intrinsic order;
\State Finetune the complete pruned network;
\State Add the finetuned network $\Phi'$ into $A$;
\State Set $\Phi_0 = \Phi'$; 
\State $i  = i  + 1$
\EndWhile
\State \textbf{return} the pruned networks in $A$
\end{algorithmic}
\end{algorithm}

\subsection{Details of the EA in Each Group}

\begin{figure*}[htbp]  \centering  \includegraphics[width=\textwidth]{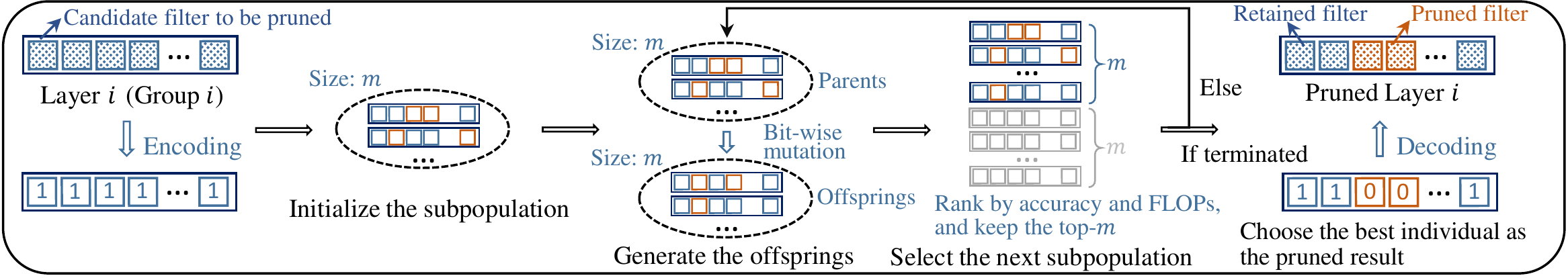}
\vspace{-1.5em}
\caption{Illustration of the EA in each group of CCEP. Given the candidate filters of the $i$th layer to be pruned, it employs an iterative evolutionary optimization process to search for an optimal pruned layer in terms of accuracy and FLOPs. Six important steps are involved: encoding, initialization, offspring generation, environmental selection, final individual selection, and decoding.}   \label{FIG:EA}\vspace{-0.8em} \end{figure*}

Generally, the EA in each group follows the typical evolutionary framework. It starts by initializing a subpopulation for the current group, and then evolves the subpopulation iteratively by generating and evaluating new individuals, and selecting better individuals to remain in the next generation. When terminating, the EA selects an individual from the final subpopulation and outcomes the corresponding pruned layer. The whole process is shown in Algorithm~\ref{alg:EA} and Figure~\ref{FIG:EA}. 

\begin{algorithm}[tb]
\caption{EA in each group}
\label{alg:EA}
\textbf{Input}: The $i$th layer, population size $m$, data set $D_s$, mutation rates $p_1, p_2$, ratio bound $r$, maximum generation $G$\\
\textbf{Output}: A pruned layer\\
\textbf{Process}:
\begin{algorithmic}[1] 
\State Initialize a subpopulation $P$ with $I_0$ and $m-1$ individuals generated from $I_0$ by bit-wise mutation with $p_1$ and $r$; 
\State Let $j = 0$; 
\While{$j$ \textless \  $G$}
\State Generate $m$ offspring individuals, each of which is
\Statex \quad \ \ generated by randomly selecting a parent individual
\Statex \quad \ \ from $P$ and conducting bit-wise mutation with $p_2$ and
\Statex \quad \ \ $r$ on the parent individual; 
\State Calculate the accuracy of the $m$ individuals on $D_s$ as
\Statex \quad \ \ well as their FLOPs; 
\State Set $Q$ as the union of $P$ and the $m$ individuals; 
\State Rank the individuals in $Q$ by accuracy and FLOPs; 
\State Select the top $m$ individuals. Use them to update $P$; 
\State $j  = j  + 1$
\EndWhile
\State Conduct the final individual selection;
\State \textbf{return} the pruned layer w.r.t. the selected individual
\end{algorithmic}
\end{algorithm}

Concretely, the subpopulation is initialized with $m$ individuals in line~1 of Algorithm~\ref{alg:EA}. The first individual is created by the vector with all 1s, denoted as $I_0$, to encourage a conservative pruning. The other $m-1$ individuals are generated by applying the following bit-wise mutation operator with a mutation rate $p_1$ on $I_0$. Specifically, given an individual, the bit-wise mutation operator flips each bit of the individual with some probability (called mutation rate) independently until all bits have been tried or an upper bound $r$ on the pruning ratio that limits the maximum number of pruned filters has been met. That is, the number of 0s (i.e, the number of pruned filters) of the generated individual should be no larger than $|I_0| \times r$, where $|I_0|$ is the number of filters contained by the $i$th layer of the current base network, preventing the pruning from being too violent. A detailed explanation of the bit-wise mutation operator is provided in the appendix.

\begin{figure}[tp]   \flushleft  \includegraphics[width=8.5cm]{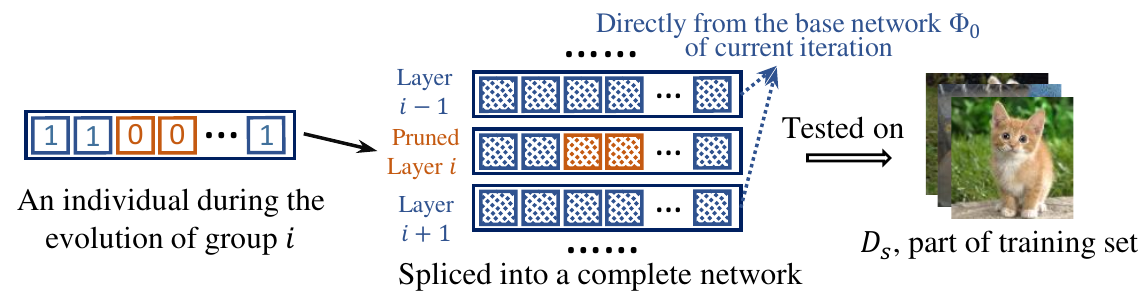}\vspace{-0.4em}
\caption{Illustration of individual splicing for evaluation in the EA of group $i$, where an individual corresponding to a pruned $i$th layer is spliced with other layers from the current base network to form a complete network, and then evaluation can be conducted as usual.}  \label{FIG:Evaluation}\vspace{-1.1em} \end{figure}

In each generation of Algorithm~\ref{alg:EA}, an offspring individual is generated by applying the bit-wise mutation operator with a mutation rate $p_2$ on a parent individual randomly selected from $P$, and this process is repeated independently to generate $m$ offspring individuals in line~4. When evaluating an offspring individual in line~5, which corresponds to a single pruned layer, a complete network is first formed by obtaining other layers from the base network of the current iteration and splicing them with the current pruned layer in their intrinsic order, as shown in Figure~\ref{FIG:Evaluation}. The resultant complete network is evaluated in terms of accuracy, which is conducted on a small randomly sampled data set $D_s$ to improve efficiency, and FLOPs. The results are assigned back to the individual under evaluation. After that, the $m$ offspring individuals and the individuals in the current subpopulation $P$ will be merged and ranked in lines~6--7, and the top $m$ individuals will be selected to form the new subpopulation for the next generation in line~8. For the ranking, all individuals are ranked by their accuracy descendingly, and for the individuals with the same accuracy, the one with fewer filters is ranked higher. Note that to reproduce offspring individuals, only the simple bit-wise mutation operator is used, but the experimental results in Section~\ref{sec-exp} have shown that it is good enough. Using more sophisticated operators (e.g., crossover operators) is expected to further improve the performance.

After running $G$ generations, the final individual selection will be carried out. Two strategies are considered. One is to always select the top-ranked individual in the final subpopulation, called strategy $Sel_{a}$. The other strategy called $Sel_{b}$, is to select the top-ranked one except the individual with all 1s. That is, if the top-ranked individual in the final subpopulation has all bits equal to 1 (i.e., does not prune any filter), the strategy $Sel_{b}$ will select the runner-up one. $Sel_{b}$ can be used to speed up the whole pruning process while mildly sacrificing accuracy, which is useful in complex problems. The influence of these two strategies will be examined in the experiments.

\section{Experiments}\label{sec-exp}

The experiments are conducted from two aspects. The first is to compare CCEP with existing pruning methods, including not only state-of-the-art pruning methods which require experts to design good criteria or auxiliary modules, but also automatic EA-based pruning methods. The second is to visualize the architecture of the pruned networks, and examine the effectiveness of introducing the pruning ratio bound $r$ in mutation and the influence of hyper-parameters of CCEP. 

Two classic data sets CIFAR10~\cite{krizhevsky2009learning} and ImageNet~\cite{ILSVRC15} for image classification are used for the examination. Two typical networks with different architectures are examined, i.e., VGG~\cite{DBLP:journals/corr/SimonyanZ14a} and ResNet~\cite{DBLP:conf/cvpr/HeZRS16}. Following the common settings, the pruning is on all convolution layers for VGG, while it is on the first convolution layer of the normal residual blocks and the first and second convolution layers of the bottleneck blocks for ResNet. 

The methods for comparison include six typical criteria-based pruning methods: L1~\cite{DBLP:conf/iclr/0022KDSG17}, SFP~\cite{DBLP:conf/ijcai/HeKDFY18}, FPGM~\cite{DBLP:conf/cvpr/HeLWHY19}, ThiNet~\cite{DBLP:journals/pami/LuoZZXWL19}, HRank~\cite{lin2020hrank} and ManiDP~\cite{DBLP:conf/cvpr/Tang0XD0T021}; seven learning-based methods: AMC~\cite{DBLP:conf/eccv/HeLLWLH18}, GAL~\cite{DBLP:conf/cvpr/LinJYZCYHD19}, MetaPruning~\cite{DBLP:conf/iccv/LiuMZG0C019}, LFPC~\cite{DBLP:conf/cvpr/HeDLZZ020},  Pscratch~\cite{DBLP:conf/aaai/WangZXZSZH20}, BCNet~\cite{su2021bcnet} and GReg~\cite{DBLP:conf/iclr/WangQZ021}; and two EA-based methods: DeepPruningES~\cite{DBLP:journals/isci/FernandesY21} and ESNB~\cite{2021Evolutionary}. All the results of the pruning ratio of FLOPs and accuracy are obtained directly from their original reports. 

\subsection{Experiments on CIFAR-10}

CIFAR-10 has 50K images in the training set and 10K images in the test set, with the size of 32$\times$32 for 10 categories. ResNet-56, ResNet-110 and VGG-16 are tested, where the common variant of VGG-16~\cite{DBLP:conf/iclr/0022KDSG17} for CIFAR-10 is used here. The settings of CCEP are described as follows. It runs for 12 iterations, i.e., $T=12$ in Algorithm~\ref{alg:CCEP}. For the EA (i.e., Algorithm~\ref{alg:EA}) in each group, the population size $m$ is 5, the mutation rates $p_1=0.05$ and $p_2=0.1$, the ratio bound $r$ is 0.1, the maximum generation $G$ is 10, and $20\%$ of the training set is used for accuracy evaluation. For the final individual selection in line~11 of Algorithm~\ref{alg:EA}, the strategy $Sel_{a}$ is used. The setting of finetuning in line~7 of Algorithm~\ref{alg:CCEP} is provided in the appendix. We run CCEP independently for 10 times with different random seeds. 

\begin{figure}[h!]
\centering
	\includegraphics[width=\linewidth,keepaspectratio]{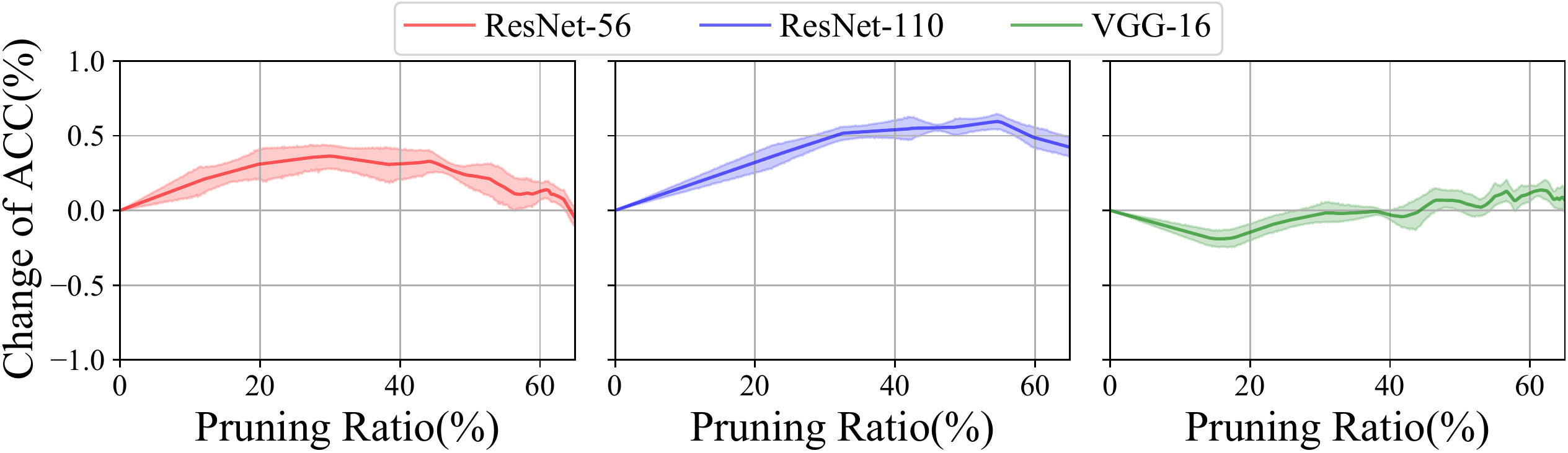}\vspace{-0.6em}
\caption{Repetition test of CCEP on CIFAR-10.}
\label{stability}\vspace{-1em}
\end{figure}

The comparison results in terms of the accuracy drop and pruning ratio of FLOPs are shown in Table~\ref{result-cifar10}. The algorithms are ranked by their pruning ratios. For CCEP, we only present the solution with a pruning ratio similar to the comparison methods in Table~\ref{result-cifar10} from the final generated archive $A$. Generally, CCEP not only outperforms state-of-the-art pruning methods which require experts to design good criteria or auxiliary modules, but also automatic EA-based pruning methods. In most cases, CCEP can achieve both larger pruning ratio and smaller accuracy drop. As for the comparison with DeepPruningES, CCEP shows an obvious superiority on accuracy drop with similar pruning ratios. Additionally, we present the curves of pruning process of CCEP during the 10 independent runs. As shown in Figure~\ref{stability}, the solid line is the mean value of the 10 independent runs, and the shadow area represents the $95\%$ confidence interval. The results imply a good stability of CCEP.\vspace{-0.5em}

\begin{table}[t!]
\small
\centering
\begin{tabular}{@{}lcccc@{}}
\toprule
Method & Architecture & \begin{tabular}[c]{@{}c@{}}Base\\ACC(\%)\end{tabular} & \begin{tabular}[c]{@{}c@{}}ACC $\downarrow$\\(\%)\end{tabular} & \begin{tabular}[c]{@{}c@{}}FLOPs $\downarrow$\\(\%)\end{tabular} \\ \midrule
Hrank & \multirow{11}{*}{ResNet-56} & 93.26 & 0.09 & 50.00 \\
AMC &  & 92.80 & 0.90 & 50.00 \\
Pscratch & & 93.23 & 0.18 & 50.00 \\
FPGM & & 93.59 & 0.33 & 52.30 \\
SFP & & 93.59 & 0.24 & 52.60 \\
ESNB & & 93.59 & 0.62 & 52.60 \\
LFPC & & 94.37 & 0.35 & 52.90 \\
GAL-0.8 & & 93.26 & 1.68 & 60.20 \\
ManiDP & & 93.70 & 0.06 & 62.40 \\
\textbf{CCEP} & & 93.48 & \textbf{-0.24} & \textbf{63.42} \\
DeepPruningES & & 93.37 & 2.65 & 66.23 \\ \midrule
ESNB & \multirow{7}{*}{ResNet-110} & 93.25 & -0.12 & 25.19 \\
SFP & & 93.68 & -0.18 & 40.80 \\
GAL-0.5 & & 93.50 & 0.76 & 48.50 \\
FPGM & & 93.68 & -0.17 & 52.30 \\
Hrank & & 93.50 & 0.14 & 58.20 \\
DeepPruningES & & 93.80 & 2.46 & 64.84 \\
\textbf{CCEP} & & 93.68 & \textbf{-0.22} & \textbf{67.09} \\ \midrule
L1 & \multirow{6}{*}{VGG-16} & 93.25 & -0.15 & 34.20 \\
GAL-0.05 & & 93.96 & 0.19 & 39.60 \\
BCNet & & 93.99 & -0.37 & 50.63 \\
HRank & & 93.96 & 0.53 & 53.50 \\
\textbf{CCEP} & & 93.71 & \textbf{-0.19} & \textbf{63.20} \\
DeepPruningES & & 93.94 & 2.98 & 65.49 \\ \bottomrule
\end{tabular}\vspace{-0.3em}
\caption{Comparison in terms of accuracy drop and pruning ratio on CIFAR-10. The algorithms are listed in ascending order of the pruning ratio. The results of our algorithm CCEP are shown in bold.}
\label{result-cifar10}\vspace{-0.8em}
\end{table}

\subsection{Experiments on ImageNet}

ImageNet contains 1.28M images in the training set and 50K in the validation set, for 1K classes. The size of each image is 224$\times$224. ResNet-34 and ResNet-50 are examined. Since ImageNet is much complex than CIFAR-10, a slightly larger mutation rate and pruning ratio bound (i.e., $p_1=0.1$ and $r=0.15$) are used to improve the pruning speed, and also only $1\%$ of the training set is used for accuracy evaluation. The other settings of CCEP are same as that on CIFAR-10. 

\begin{figure}[t!]  \centering
\includegraphics[width=6.5cm,keepaspectratio]{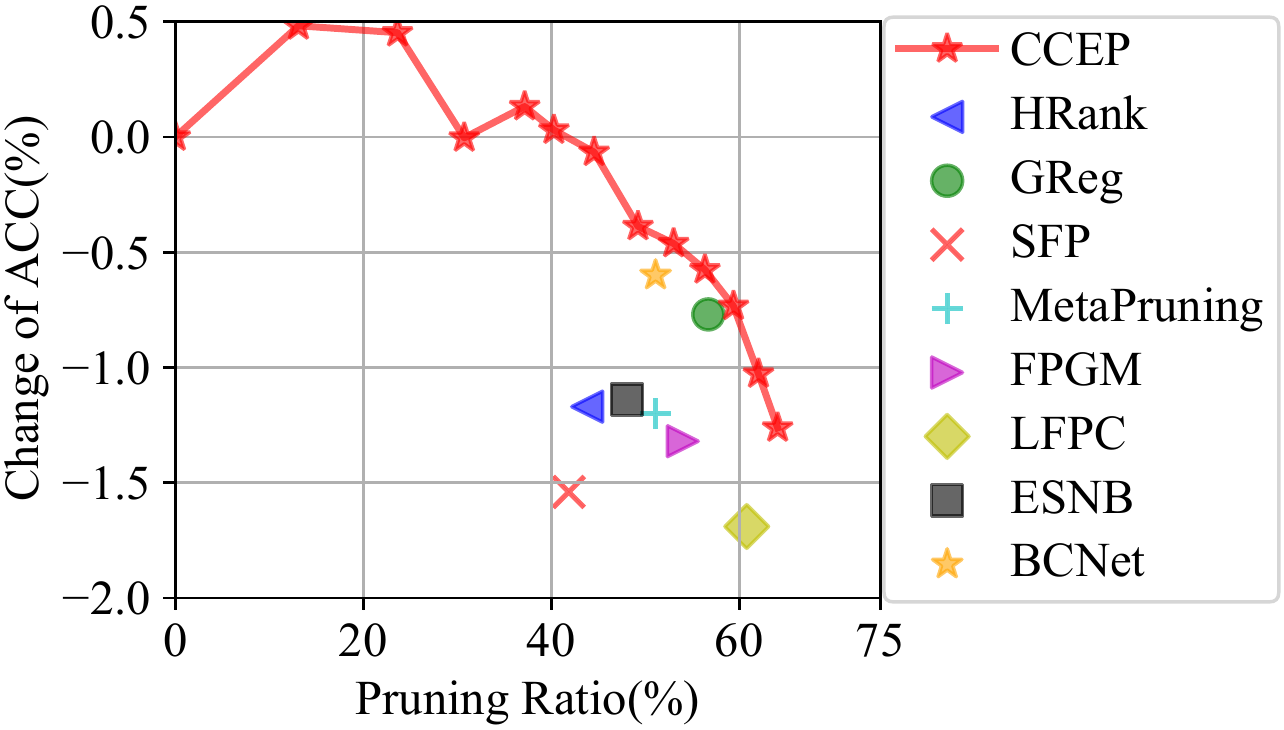} \vspace{-0.5em} \caption{Pruning comparison for ResNet-50 on ImageNet.} \vspace{-0.8em}
\label{ResNet50-imagenet}
\end{figure}

The results are shown in Table~\ref{result-imagenet}, where the algorithms are ranked by their pruning ratios. For CCEP, we select three solutions, which have pruning ratios similar to that in Table~\ref{result-imagenet}, from the final archive $A$ generated in line~12 of Algorithm~\ref{alg:CCEP}. Generally, CCEP surpasses its counterparts at different levels of pruning ratio with less drop of accuracy. To the best of our knowledge, this is the first time that an EA-based filter pruning method has been examined on ImageNet. The superiority of CCEP on such a large data set provides a more comprehensive demonstration of its effectiveness for neural network pruning. Besides, all the pruned networks (i.e., the archive $A$ in Algorithm~\ref{alg:CCEP}) generated during the pruning process of CCEP on ResNet-50 are shown in Figure~\ref{ResNet50-imagenet}. On one hand, CCEP can achieve a set of pruned networks with different pruning ratios in a single run, providing a better flexibility for users. On the other hand, the curve implies the ability of CCEP in finding a good tradeoff between accuracy and pruning ratio. That is, at the beginning, the accuracy of the pruned network can increase as the pruning goes on, implying that the pruned redundant filters might bring noise. In the later stages, the accuracy starts to fall as too many filters are pruned and the representation ability might be damaged.

\begin{table*}[t!]
\small
\centering
\begin{tabular}{@{}llccccccc@{}}
\toprule
Architecture & Method & \begin{tabular}[c]{@{}c@{}}Base \\ Top-1(\%)\end{tabular} & \begin{tabular}[c]{@{}c@{}}Pruned \\ Top-1(\%)\end{tabular} & \begin{tabular}[c]{@{}c@{}}Base \\ Top-5(\%)\end{tabular} & \begin{tabular}[c]{@{}c@{}}Pruned \\ Top-5(\%)\end{tabular} & Top-1 $\downarrow$& Top-5 $\downarrow$ & FLOPs $\downarrow$(\%) \\ \midrule
\multirow{5}{*}{ResNet-34} & GReg & 73.31 & 73.61 & - & - & -0.30 & -  & 24.24 \\
 & \textbf{CCEP-1} & \textbf{73.30} & \textbf{73.64} & \textbf{91.42} & \textbf{91.51} & \textbf{-0.34} & \textbf{0.11}  & \textbf{25.44} \\
 & FPGM & 73.92 & 72.63 & 91.62 & 91.08 & 1.29 & 0.54  & 30.00 \\
 & SFP & 73.92 & 71.83 & 91.62 & 90.33 & 2.09 & 1.29 & 41.10 \\
 & \textbf{CCEP-2} & \textbf{73.30} & \textbf{72.67} & \textbf{91.42} & \textbf{90.90} & \textbf{0.63} & \textbf{0.72}  & \textbf{42.10} \\ \midrule
\multirow{12}{*}{ResNet-50} & SFP & 76.15 & 74.61 & 92.87 & 92.06 & 1.54 & 0.81 & 41.86 \\
 & \textbf{CCEP-1} & \textbf{76.13} & \textbf{76.06} & \textbf{92.86} & \textbf{92.81} & \textbf{0.07} & \textbf{0.05}  & \textbf{44.56} \\
 & ESNB & 77.27 & 76.13 & - & - & 1.14 & -  & 48.06 \\
 & MetaPruning & 76.60 & 75.40 & - & - & 1.20 & -  & 51.10 \\
 & BCNet & 77.50 & 76.90 & - & 93.30 & 0.60 & -  & 51.10 \\
 & FPGM & 76.15 & 74.83 & 92.87 & 92.32 & 1.32 & 0.55  & 54.00 \\
 & ThiNet & 75.30 & 72.03 & 92.20 & 90.99 & 3.27 & 1.21  & 55.75 \\
 & \textbf{CCEP-2} & \textbf{76.13} & \textbf{75.55} & \textbf{92.86} & \textbf{92.63} & \textbf{0.58} & \textbf{0.23} & \textbf{56.35} \\
 & GReg & 76.13 & 75.36 & - & - & 0.77 & - & 56.71 \\
 & LFPC & 76.13 & 74.46 & 92.87 & 92.32 & 1.67 & 0.55  & 60.80 \\
 & Hrank & 76.15 & 71.98 & 92.87 & 91.01 & 4.17 & 1.86  & 61.12 \\
 & \textbf{CCEP-3} & \textbf{76.13} & \textbf{74.87} & \textbf{92.86} & \textbf{92.35} & \textbf{1.26} & \textbf{0.51}  & \textbf{64.09} \\ \bottomrule
\end{tabular}\vspace{-0.5em}
\caption{Comparison in terms of accuracy drop and pruning ratio on ImageNet. The algorithms are listed in ascending order of the pruning ratio. The results of our algorithm CCEP are shown in bold. The `-' means that the corresponding result is not provided in its original paper.}
\label{result-imagenet}\vspace{-0.6em}
\end{table*}

\begin{figure*}[t!]
\centering
	\subfloat[ResNet-56]{\includegraphics[width = 0.25\textwidth]{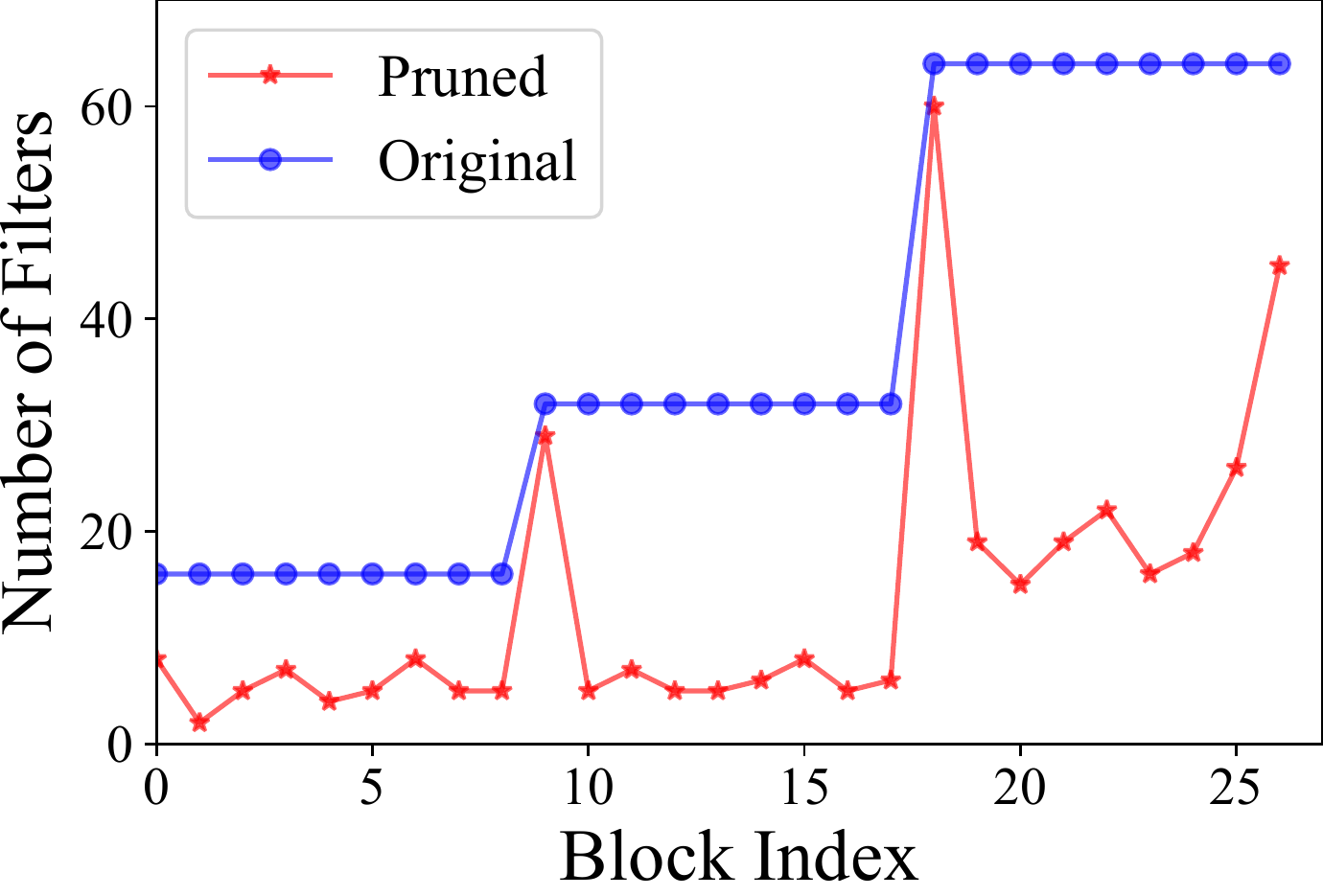}}
	\hspace{8mm}
	\subfloat[ResNet-110]{\includegraphics[width = 0.25\textwidth]{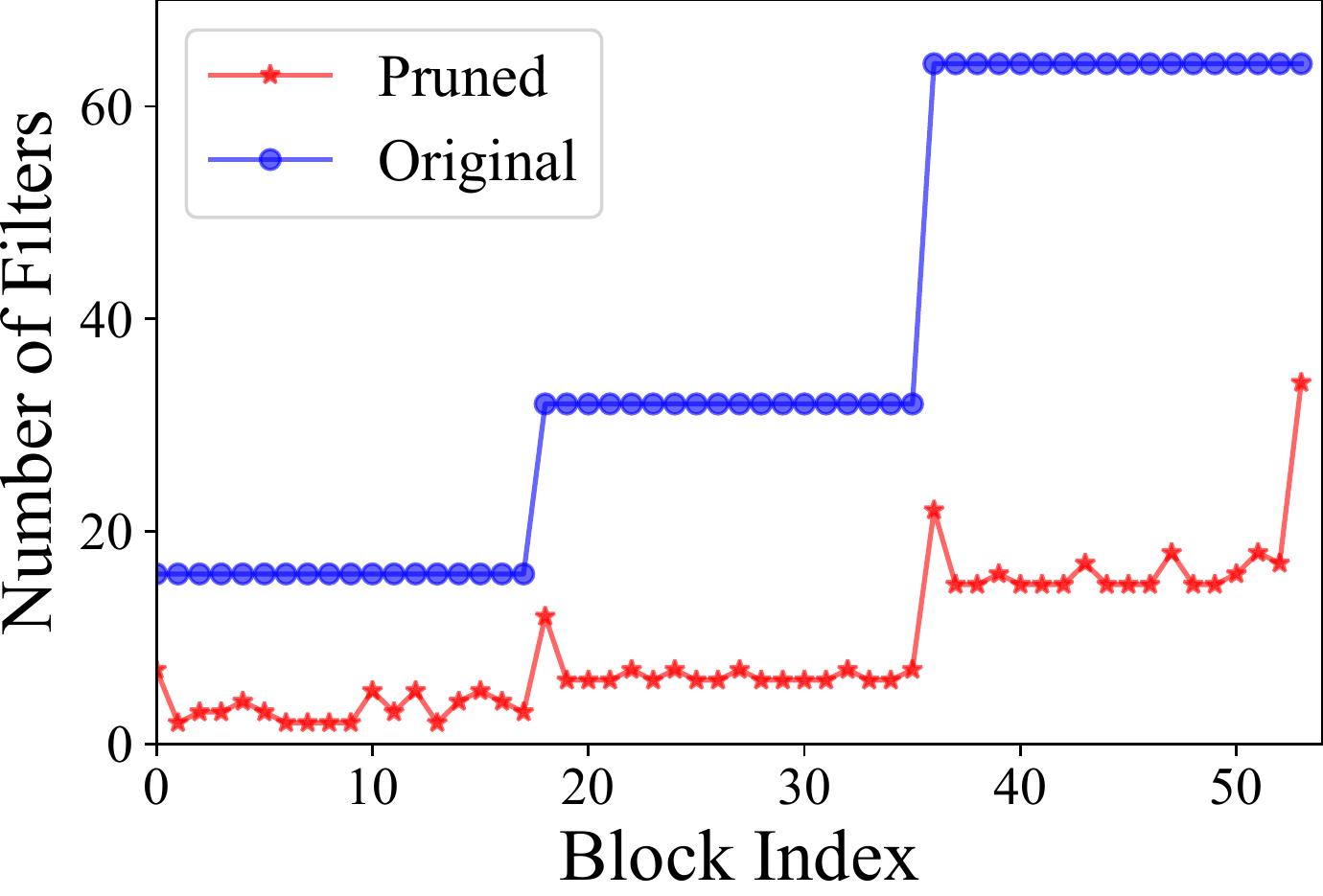}} 
	\hspace{8mm}
	\subfloat[VGG-16]{\includegraphics[width = 0.25\textwidth]{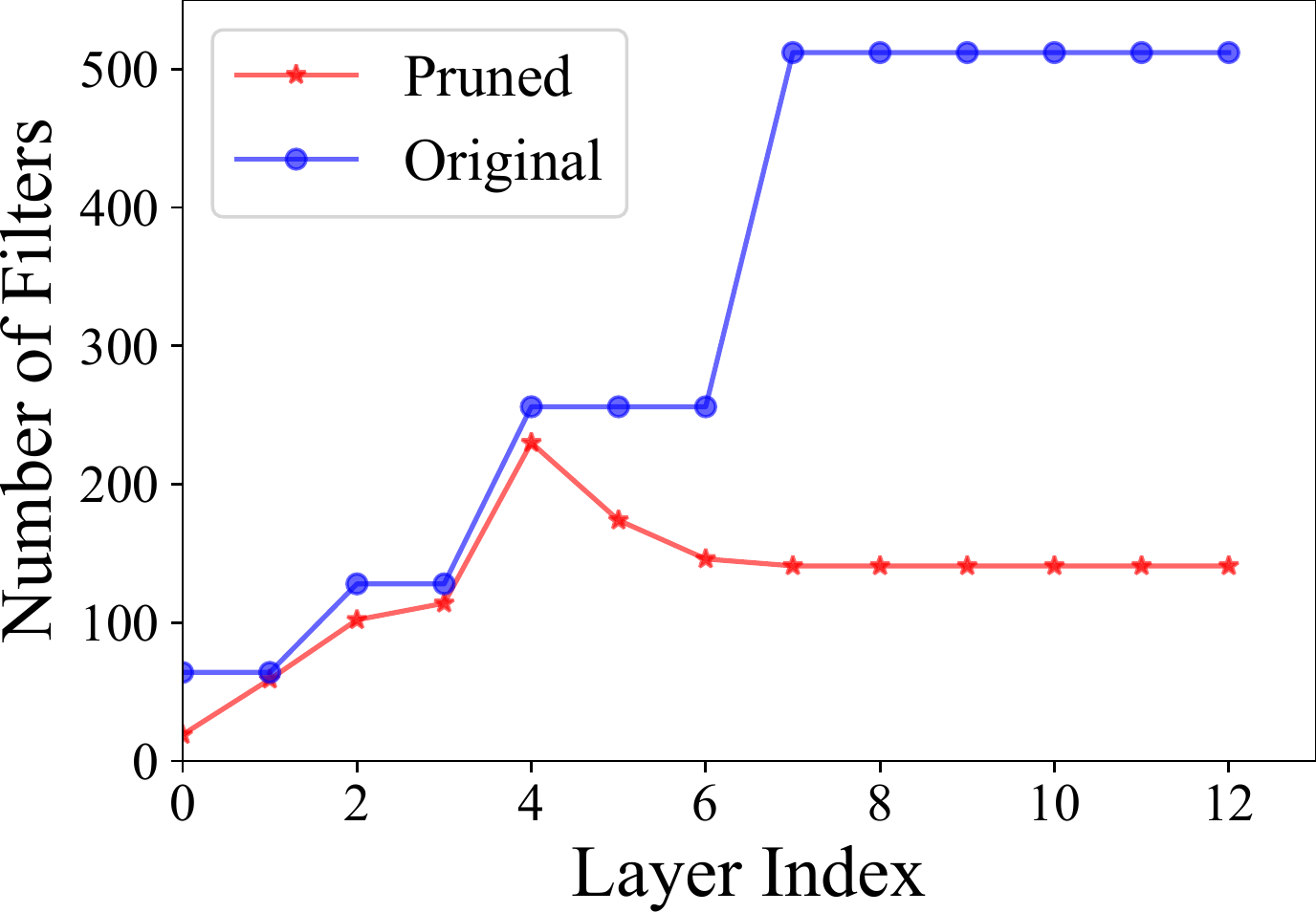}}\vspace{-0.7em}
\caption{Visualization of the original networks and the pruned networks obtained by CCEP on CIFAR-10.}
\label{shape}\vspace{-1.1em}
\end{figure*}

\subsection{Further Studies and Discussion}

Figure~\ref{shape} visualizes the architectures (i.e., the number of filters in each block or layer) of the pruned networks on CIFAR-10, obtained by CCEP. We can observe that on ResNet-56 and ResNet-110, the filters around the expansion of channels are less pruned than others; while on VGG-16, the filters at the tail part tend to be pruned. These results imply that CCEP can achieve different degrees of redundancy in different layers automatically, which may be difficult to be captured manually.

We have introduced the parameter of ratio bound $r$ in mutation, to limit the ratio of pruned filters in each iteration of CCEP. Its effect is examined by comparing CCEP with and without $r$, which are run five times independently for pruning ResNet-56 on CIFAR-10. Figure~\ref{FIG:ratio_bound} shows the average accuracy change and pruning ratio of their generated pruned networks. As expected, using $r$ brings significantly better accuracy by preventing overly aggressive pruning, which would make it hard or even fail to recover the network accuracy. 

\begin{figure}[b!]   \flushleft 
\includegraphics[width=8.5cm]{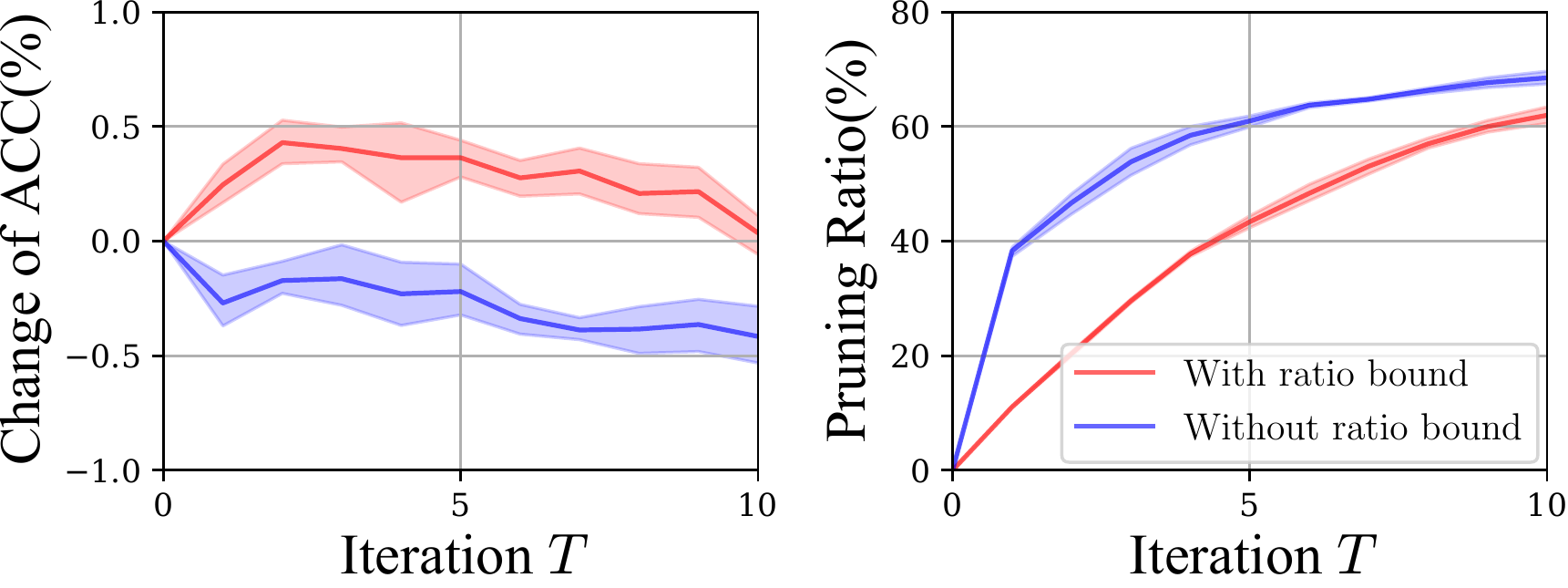}\vspace{-0.8em}
\caption{Comparison of CCEP with and without the ratio bound $r$ for pruning ResNet-56 on CIFAR-10.}  \label{FIG:ratio_bound}\vspace{-0.7em} \end{figure}

The influence of hyper-parameters (including the population size $m$, maximum generation $G$, mutation rate $p_1$, ratio bound $r$ and final individual selection strategy) of CCEP is examined by pruning ResNet-56 on CIFAR-10, and provided in the appendix. Generally, a larger $G$, or a larger $p_1$ and $r$ would lead to a faster pruning process but at an expense of accuracy. For the final individual selection strategy, $Sel_a$ achieves better accuracy while $Sel_b$ shows superiority on efficiency especially on the complex data set. This is intuitive, as a faster pruning process is usually more violent. These results can guide the settings of CCEP to some extent in practice, and the settings used in our experiments are to make a trade-off between the pruning speed and accuracy.

\section{Conclusion}

This paper proposes an automatic neural network pruning algorithm CCEP. It applies cooperative coevolution to tackle the large search space of neural network pruning, where the layered characteristic of typical neural networks is utilized to tackle the non-trivial separation issue in cooperative coevolution. Experiments show the competitive performance of CCEP, compared with a number of state-of-the-art pruning methods. Besides, the framework of cooperative coevolution makes CCEP highly flexible and easily parallelizable. In the future, we will try to incorporate more advanced EA operators to further improve the performance of CCEP.

\begin{small}
    \bibliographystyle{named}
    \bibliography{ijcai22-ccep}
\end{small}

\clearpage
\section{Appendix}
The appendix is to provide the detailed settings of finetuning a pruned network during the running of CCEP, detailed description of the bit-wise mutation operator, and the stuides on the influence of hyper-paramters of CCEP, which are omitted in the main body of this paper due to space limitation. 

\subsection{Detailed Settings of Finetuning}

As shown in Algorithm~1, CCEP employs an iterative prune-and-finetune strategy. In our experiments, the settings for finetuning a network on CIFAR-10 a and ImageNet are shown in Table~\ref{finetune}, which are  the common settings used in the literature. On CIFAR-10, the SGD optimizer is used, and its learning rate is initially set as 0.1, multiplied by 0.1 after the milestone of 50 epochs. The total number of epochs is 100. The momentum is 0.9 and the weight decay is 0.0001. The batch size for training is 128. On ImageNet, the learning rate of the SGD optimizer is initially set as 0.01, multiplied by 0.1 at epoch 20, 40 and 50, respectively. The total number of epochs is 60. The momentum is 0.9 and the weight decay is 0.0001. The batch size for training is 256. Note that since the complexity and size of the two data sets differ greatly, the proper parameters of finetuning are also different.

\begin{table}[h!]
\centering
\small
\setcounter{table}{2}
\begin{tabular}{@{}cccc@{}}
\toprule
 & CIFAR-10 & ImageNet & \\ \midrule
Optimizer & SGD & SGD\\
Initial learning rate & 0.1 & 0.01 \\
Epochs & 100 & 60 \\
Milestones($\times$0.1) & [50] & [20,40,50] \\
Weight decay & 0.0001 & 0.0001 \\
Momentum & 0.9 & 0.9 \\
Batch size & 128 & 256 \\ \bottomrule
\end{tabular}
\caption{Settings of the finetuning process.}
\label{finetune}
\end{table}

\subsection{Influence of Hyper-parameters of CCEP}

CCEP has several hyper-parameters, and we study their influence on the performance of CCEP by the experiments of pruning ResNet-56 on CIFAR-10. Particularly, we consider the population size $m$, maximum generation $G$, mutation rate $p_1$, ratio bound $r$, and the final individual selection strategy. For CCEP with each hyper-parameter configuration, the maximum iteration $T$ is set as 10, and we repeat its run for five times independently with different random seeds and report the average results. 

The results may guide the hyper-parameter settings of CCEP to some extent in practice, and the settings used in our experiments are to make a trade-off between the pruning speed and accuracy. As shown below, we actually did not pay much effort on tuning the hyper-parameters and only tested a very limited number of settings for each hyper-parameter, which, however, has been already able to lead to a good performance of CCEP as reported in the original paper. It is believed that further improvements on the performance of CCEP can be achieved by more careful tuning.

\subsubsection{Population Size}

Generally, a larger population size may help EAs escape from local optima, but also requires more evaluations in one generation. Considering the time-consuming training of deep CNNs in evaluation, we test $m$ with the values of $\{3,5,7,9\}$. The results in Figure~\ref{FIG:population_size} show that the impact of $m$ within this range on the performance CCEP is small. We set $m$ to 5 in our experiments.

\begin{figure}[h!]   \flushleft 
\setcounter{figure}{7}
\includegraphics[width=8.5cm]{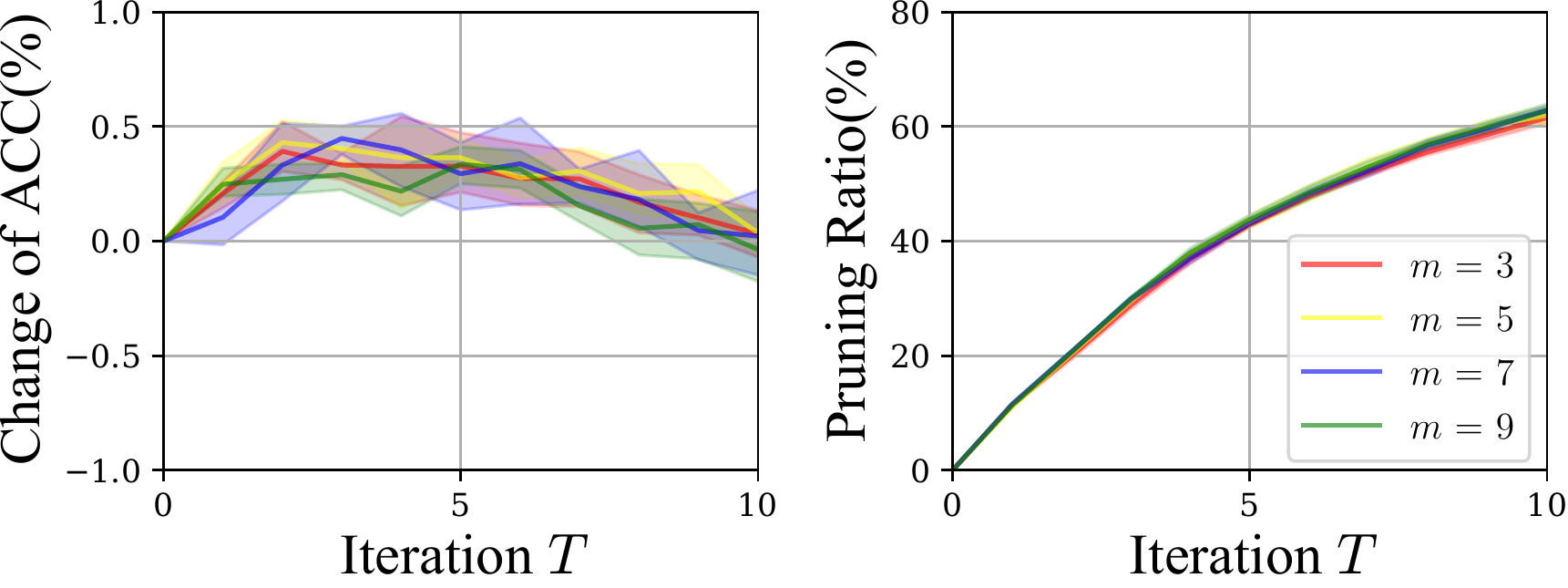} 
\caption{Influence of the population size $m$.}  \label{FIG:population_size} \end{figure}

\subsubsection{Maximum Generation}

\begin{figure}[h!]   \flushleft 
\includegraphics[width=8.5cm]{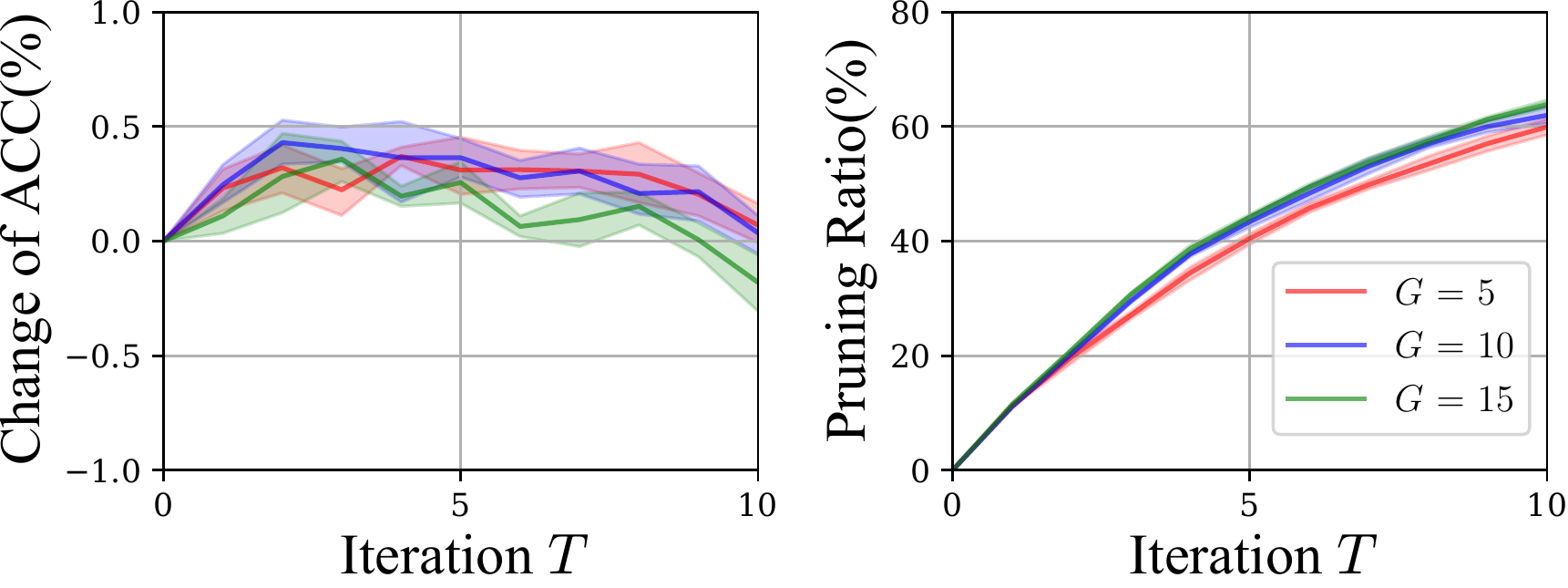} 
\caption{Influence of the maximum generation $G$.}  \label{FIG:T} \end{figure}

For the EA in each group, we test its maximum generation $G$ with the values of $\{5,10,15\}$. As shown in Figure~\ref{FIG:T}, a larger $G$ will lead to a faster pruning process but at an expense of accuracy. This is expected, because a larger $G$ implies a more aggressive pruning by the EA in each group, which could make the recovery of network accuracy difficult. To make a trade-off between the pruning speed and accuracy, the maximum generation $G$ is set to 10 in our experiments.

\subsubsection{Mutation Rate and Ratio Bound}

\begin{figure}[h!]   \flushleft 
\includegraphics[width=8.5cm]{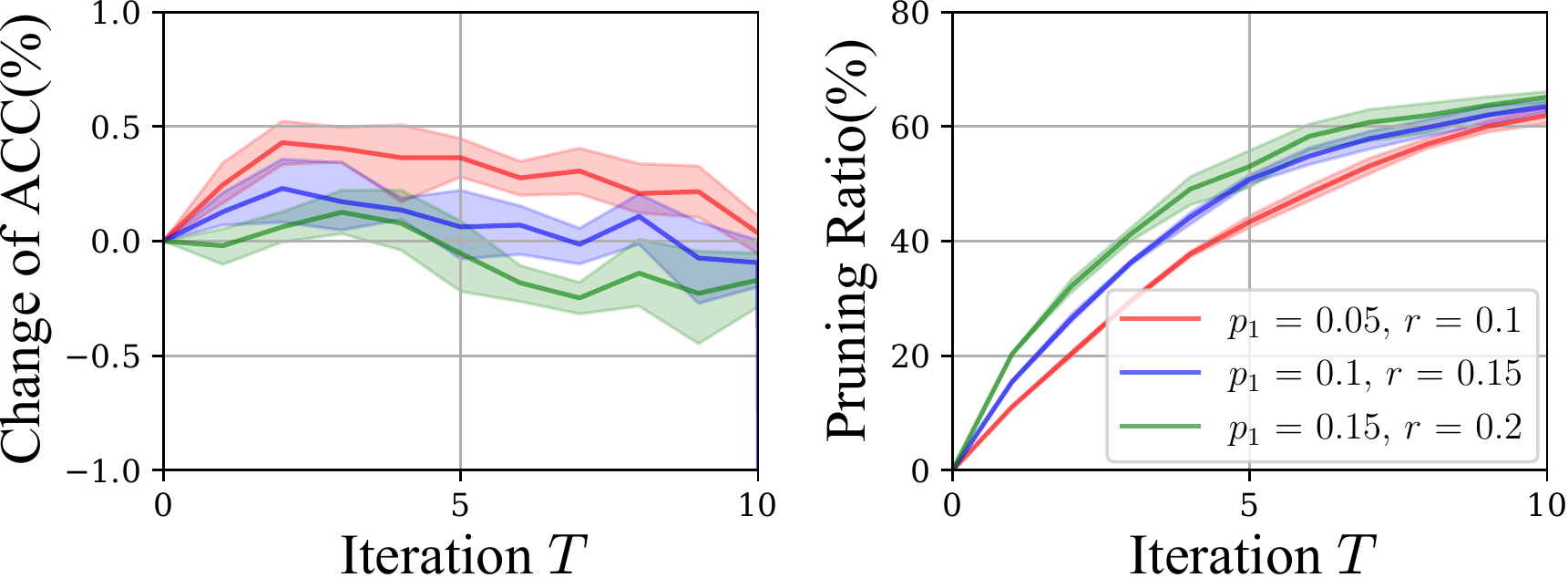} 
\caption{Influence of the mutation rate $p_1$ and ratio bound $r$.}  \label{FIG:mutation} \end{figure}

The mutation rate $p_1$ and the ratio bound $r$ are examined together, because they are strongly correlated. That is, when $p_1$ is large, more filters may be removed, and thus a larger pruning ratio bound $r$ is needed. If $r$ is small, setting a large $p_1$ would be meaningless. Three different settings are tested, i.e., $p_1=0.05 \wedge r=0.1$, $p_1=0.1 \wedge r=0.15$ and $p_1=0.15 \wedge r=0.2$. The results in Figure~\ref{FIG:mutation} show that a larger setting leads to a faster pruning process, but at an expense of accuracy. Thus, to make a trade-off between the pruning speed and accuracy, the setting of $p_1=0.05 \wedge r=0.1$ is used for the small data set CIFAR-10 while $p_1=0.1 \wedge r=0.15$ is used for the large data set ImageNet.

\subsubsection{Final Individual Selection Strategy}

\begin{figure}[h!]   \flushleft 
\includegraphics[width=8.5cm]{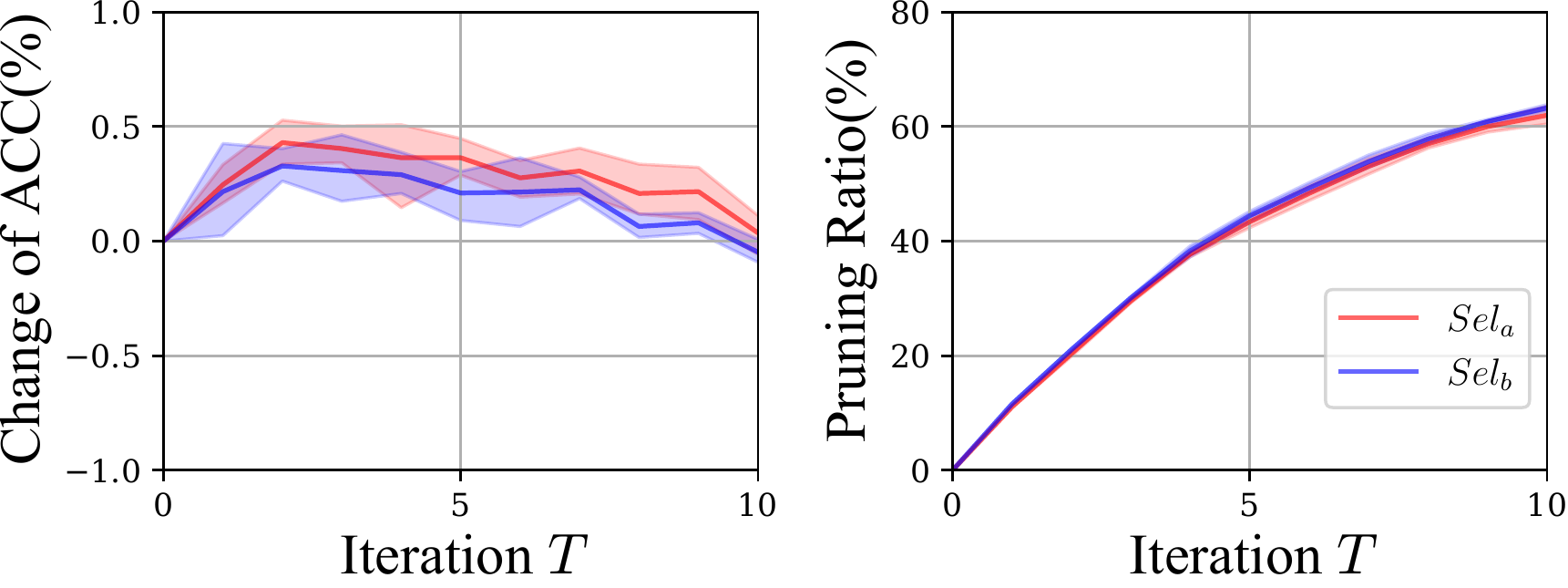} 
\caption{Influence of the final individual selection strategy for pruning ResNet-56 on CIFAR-10.}  \label{FIG:selection_strategy_Cifar10} \end{figure}

For the EA in each group, we have provided two final individual selection strategies, i.e., $Sel_a$ and $Sel_b$, where $Sel_a$ always selects the best individual from the final subpopulation, and $Sel_b$ selects the best one except the individual with all 1s. The results for pruning ResNet-56 on CIFAR-10 are shown in Figure~\ref{FIG:selection_strategy_Cifar10}. As expected, $sel_a$ achieves better accuracy while $sel_b$ shows superiority on efficiency. This is because $sel_b$ avoids selecting the best individual if it is the unpruned network, which may lead to a faster pruning but with an expense of accuracy.

\begin{figure}[h!]   \flushleft 
\includegraphics[width=8.5cm]{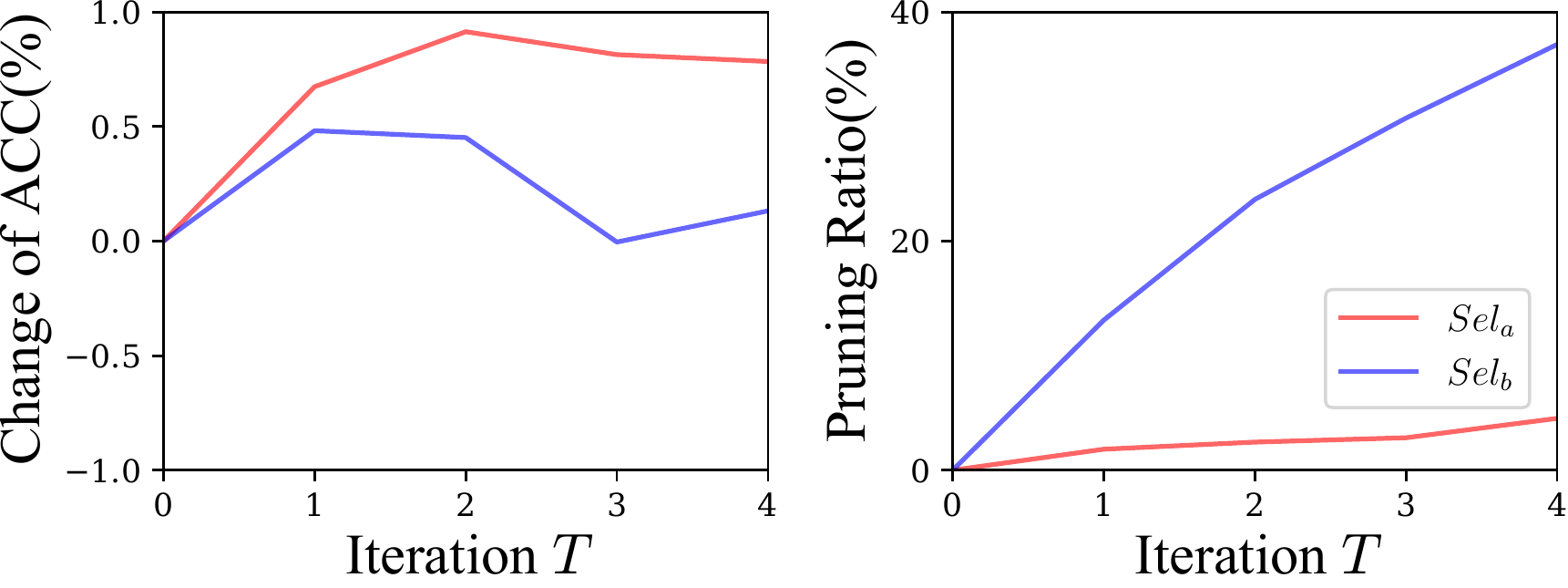} 
\caption{Influence of the final individual selection strategy for pruning ResNet-50 on ImageNet.}  \label{FIG:selection_strategy_imagenet} \end{figure}

We also compare these two strategies for pruning ResNet-50 on ImageNet, and such a phenomenon (i.e., $sel_a$ achieves better accuracy while $sel_b$ shows superiority on efficiency) is more clear, as shown in Figure~\ref{FIG:selection_strategy_imagenet}. For the more complex data set ImageNet, the network may have low redundancy, and the best individual in the final subpopulation is probably the unpruned network, leading to very different behaviors of the two strategies. In our experiments, to make a trade-off between the pruning speed and accuracy, $sel_a$ is used for the small data set CIFAR-10 with more emphasis on accuracy, while $sel_b$ is used for the large data set ImageNet with more emphasis on the pruning speed.

\subsection{Bit-wise mutation operator}
As described in subsection 3.2, the bit-wise mutation operator is used in the process of EA in each group to generate new individuals, where an upper ratio bound $r$ is introduced to limit the maximun number of pruned filters in a single iteration of CCEP, preventing the pruning from being too violent. To make it more clear, the pseudo code of the mutation operator is provided in Algorithm~\ref{alg:mutation}.

\begin{algorithm}[h]
\caption{Bit-wise mutation with a ratio bound}
\label{alg:mutation}
\textbf{Input}: A Boolean vector $\bm{s}$ with length $l$ as a parent solution, mutation rate $p$, ratio bound $r$ \\
\textbf{Output}: A Boolean vector $\bm{s}'$ as an offspring solution\\
\textbf{Process}: 
\begin{algorithmic}[1] 
\State Let $i = 1$ and $\bm{s}' = \bm{s}$;
\While {$i\leq l$} \Comment{Scan from the first bit of $\bm{s}'$.}
\State Sample a number $q$ from $[0,1]$ uniformly at random;
\If{$q<p$}  \OneLineComment{Attempt to flip the $i$th bit (denoted as $s'_i$) of $\bm{s}'$.}
\If{$s'_i = 0$}
\State $s'_i=1$ \OneLineComment{If the $i$th bit is 0, flip it to 1.}
\ElsIf{$s'_i=1$ and $|\bm{s}'|_0<l\cdot r$ }
\LeftComment{If the $i$th bit is 1 and the number}
\LineComment{of 0 do not, exceed $l \times r$,filp it to 0.}
\LineComment{Else, give up the flip.}
\State $s'_i = 0$ 
\EndIf
\EndIf
\State $i = i + 1$  
\EndWhile
\State \textbf{return} $\bm{s}'$
\end{algorithmic}
\end{algorithm}

\end{document}